%% file: main.tex
\pdfoutput=1

\documentclass[11pt]{article}

\usepackage{EMNLP2023}

\usepackage{times}
\usepackage{latexsym}

\usepackage[T1]{fontenc}

\usepackage[utf8]{inputenc}

\usepackage{microtype}

\usepackage{inconsolata}

\usepackage{amsmath,amsfonts,bm}
\usepackage{tabularx}
\usepackage{booktabs}
\usepackage{dsfont}
\usepackage{multirow}
\usepackage[inline]{enumitem}
\usepackage{color}
\usepackage{soul}
\usepackage[textsize=scriptsize]{todonotes}
\usepackage{bbm}

\title{Groundedness in Retrieval-augmented Long-form Generation:\\An Empirical Study}

\author{Alessandro Stolfo\thanks{\ \ Work partially carried out while at Oracle Labs.} \\
  ETH Z\"urich \\
  \texttt{stolfoa@ethz.ch} 
}
\begin{document}
\maketitle
\begin{abstract}
We present an empirical study of \emph{groundedness} in long-form question answering (LFQA) by retrieval-augmented  large language models (LLMs).
In particular, we evaluate whether every generated sentence is grounded in the retrieved documents or the model's pre-training data.
Across 3 datasets and 4 model families, our findings reveal that a significant fraction of generated sentences are consistently ungrounded, even when those sentences contain correct ground-truth answers.
Additionally, we examine the impacts of factors such as model size, decoding strategy, and instruction tuning on groundedness. Our results show that while larger models tend to ground their outputs more effectively, a significant portion of correct answers remains compromised by hallucinations. This study provides novel insights into the groundedness challenges in LFQA and underscores the necessity for more robust mechanisms in LLMs to mitigate the generation of ungrounded content.
\end{abstract}

\section{Introduction}

One of the most significant challenges to the safe deployment of large language models (LLMs) is their propensity to generate hallucinated content \cite{bubeck2023sparks, alkaissi2023artificial, ji2023survey}. The risk of hallucinating increases when LLMs are tasked with generating long content (i.e., more than a single sentence) \cite{shuster-etal-2021-retrieval-augmentation, maynez-etal-2020-faithfulness}. This is problematic because generating long-form text is a critical component of a number of important tasks, such as disambiguating complex topics, explicit problem decomposition and reasoning, question answering, and synthesis of information from multiple sources.

\begin{figure}[t]
    \centering
    \includegraphics[width=\columnwidth]{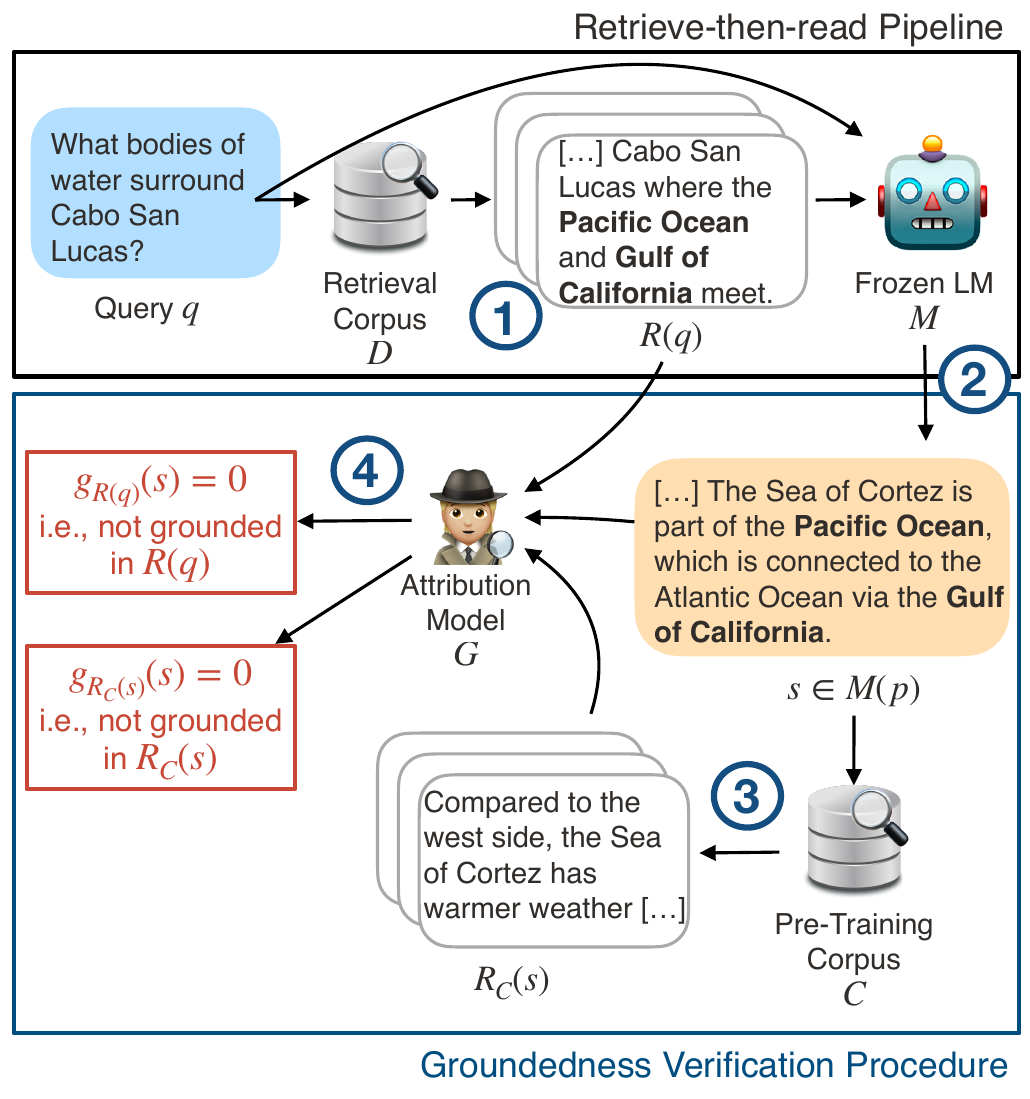}
    \caption{\textbf{Our experimental setup.}
    Using a set of retrieved documents (1), an LLM generates an answer in an LFQA setting (2). Then, the model’s pre-training corpus is searched for documents related to the generation (3). Finally, a grounding model verifies whether the model’s response is supported by any of the considered documents (4).}
  \label{fig:method}
  \vspace{-5mm}
\end{figure}

As a step towards mitigating hallucination, a number of studies have measured the \emph{groundedness} of LLM generations (for a recent survey, see \citealp{li2023survey}). In these studies, a sentence is considered to be grounded in a document if the text of the document supports the claim made in the sentence.
However, many of these efforts focus on short LLM generations (i.e., a single word or phrase) rather than long generations \cite{bohnet2022attributed}. Others rely on Google searches for exact string match as a heuristic for evaluating or improving groundedness \cite{agrawal2023language, athaluri2023exploring,gao-etal-2023-rarr}.

In this work, we focus on long-form question answering (LFQA) due to its generality and relevance. Our research addresses two central questions: \begin{enumerate*}
    \item how frequently do LLMs generate ungrounded sentences in LFQA? and,
    \item how do model size, family, pre-training recipe, and decoding style affect this rate?
\end{enumerate*} We address these questions by studying the provenance of the information contained in the model’s generations.

LFQA is typically aided by retrieval augmentation \cite[\textit{inter alia}]{karpukhin-etal-2020-dense, izacard2022few}. By harnessing external data sources, these models incorporate information pertinent to the query, reducing the likelihood of hallucinations and incorrect outputs \cite{shuster-etal-2021-retrieval-augmentation}. However, there is no guarantee that models consistently utilize the retrieved information in their outputs \cite{krishna-etal-2021-hurdles}. This setting presents additional challenges in measuring groundedness: the generated text is likely influenced by both the inference-time context (i.e., the retrieved information) and the extensive pre-training of the model. While groundedness relative to the retrieved information is well-studied, we also attempt to ground model-generated text in specific pre-training documents.

In particular, we measure whether the text generated in a retrieval-augmented fashion contains information that is \textit{grounded} in the retrieved documents or in the model's pre-training corpus (our procedure is illustrated in Figure \ref{fig:method}). We employ a groundedness-verification model \cite{honovich-etal-2022-true-evaluating} that determines whether a portion of the model's output can be attributed to a given text passage. 
We analyze four different families of pre-trained language models on three datasets.
We discover that, even when containing ground-truth answers, a significant portion of the generated sentences are not grounded in the retrieved or pre-training documents and may include fabricated claims.
This trend persists across the range of models and datasets examined.

Additionally, we study the impact of model size, decoding strategy, and instruction tuning on the rate of correct and hallucinated content. We find that the larger models are generally more adept at grounding their outputs in the given sources. However, even for the largest models analyzed (Falcon 180B; \citealp{refinedweb}), approximately 25\% of the outputs that contain ground-truth answers are not grounded.
Interestingly, we observe that instruction tuning and beam search decoding strategies contribute to a reduction in the generation of ungrounded content. These methods appear to help models better utilize training and inference-time documents, thereby mitigating the tendency to produce fabricated information.

\section{Background}
\label{sec:background}

\paragraph{Hallucination \& Factuality.}
\input{background}

\paragraph{Setting.}
We consider the task of open-domain LFQA in a few-show setting. We adopt the \textit{retrieve-then-read} paradigm, in which a language model performs question answering using passages retrieved from a corpus at inference time \cite{lewis2020retrieval, izacard-grave-2021-leveraging}. This approach, although simple, was shown to improve the few-shot performance of pre-trained LLMs on multiple QA benchmarks \cite{si2022prompting, mallen-etal-2023-trust}.

\section{Experimental Procedure}
In this section, we detail our experimental setup, including how correctness and groundedness are measured, as well as the datasets used.
We begin by defining notation.

\subsection{Notation}
Let $\mathcal{Q}$ be a collection of questions and $\mathcal{D}$ be a corpus of documents.
Consider a question $q \in \mathcal{Q}$, which is annotated with a set $\mathcal{Y}$ of ground-truth string answers. A retrieve-then-read system proceeds in 3 steps.
First, a retriever, $R: \mathcal{Q} \rightarrow \mathcal{D}$, returns a set of $k$ documents, $R(q) = \{d^{(1)}, \dots, d^{(k)}\}$. 
Second, the question and documents are combined to form a prompt, $p$.
Finally, the question-answering model, $M$, consumes the prompt and produces an answer, $M(p) = \langle s_{1}, s_{2}, \dots \rangle$, which is comprised of sentences $s_i$'s. We denote by $\mathcal{S}$ the set of all generated sentences.
In a few-shot scenario, the prompt $p$ additionally contains a set of question-documents-answer triples, which include manually annotated answers from a held-out dataset.

\subsection{Measuring Correctness}
Like previous work \cite{gao-etal-2023-enabling}, we adopt a definition of correctness based on exact match (EM).
Specifically, for a question-answer pair $(q, \mathcal{Y})$, The accuracy of a model output, $M(p)$, is computed as the fraction of elements from $\mathcal{Y}$ that are substrings of $M(p)$, i.e.,
\begin{align*}
    \mathrm{EM}(M(p), \mathcal{Y}) =\frac{|\{y \in \mathcal{Y} : \mathrm{\texttt{substr}}(y, M(p))\}|}{m},
\end{align*}
where $\mathrm{\texttt{substr}}(y, M(p)) := \mathbbm{1}\{\exists \ s \in M(p) : y \in s \}$ indicates whether $y$ is a substring of the model output, and $m = |\mathcal{Y}|$.
Concretely, in the example illustrated in Figure \ref{fig:method}, the set of ground-truth answers is $ \mathcal{Y} = \{$ \texttt{Pacific Ocean}, \texttt{Gulf of California}, \texttt{Sea of Cortez} $\}$. 
Since the model output includes the strings \texttt{Pacific Ocean} and \texttt{Gulf of California}, the accuracy of the model on this example is $\mathrm{EM}(M(p), \mathcal{Y}) = \frac{2}{3}$.
As we are interested in separately analyzing the groundedness of long-form outputs that contain correct answers and those that contain no correct answers, we refer to the set of model outputs with an exact match of $0$ as $\mathrm{EM}^0$, and all other model outputs as belonging to $\mathrm{EM}^+$.

\subsection{Measuring Groundedness}
In our work, we assume that the question-answering model, $M$, is pre-trained on a corpus, $\mathcal{C}$.
We measure the extent to which each model output is grounded in the retrieved documents, $R(q)$, as well as the training corpus, $\mathcal{C}$.
Since we focus on LFQA, we follow previous work and measure the groundedness of each sentence of each model output independently~\cite{gao-etal-2023-rarr}.

\paragraph{Groundedness in the retrieved documents.}  
Formally, let $S$ be the set of all sentences and $G: \mathcal{D} \times \mathcal{S} \rightarrow \{0,1\}$ be a grounding model, which takes a document and a sentence and outputs $1$ if the sentence is grounded in the document.
Then, a model-generated sentence, $s$, is grounded in a collection of documents, $\mathcal{Z}$, if there exists a document in $\mathcal{Z}$ that grounds $s$.
For example, for the retrieved documents, $R(q)$,
\begin{align}
\label{eq:g}
g_{R(q)}(s) =\begin{cases} 1 & \exists \ d \in R(q) : G(d, s) = 1 \\
                     0 &  \mathrm{otherwise}.
       \end{cases}
\end{align}
In words, the function $g_{R(q)}(s)$ returns 1 if at least one of the documents in $R(q)$ grounds $s$, and 0 otherwise.

\paragraph{Groundedness in the pre-training data.} 
While retrieval-augmented text-generation models have been shown to draw on the retrieved documents, they also produce sentences that are not grounded in the text provided to them during inference.
Though some of these sentences may not be grounded in any of the data ever provided to the model, others may be grounded in the model's pre-training data.
Ideally, we would compute $g_{\mathcal{C}}(s)$, i.e., whether the model-generated sentence, $s$, is grounded in \emph{any} pre-training document.
Since this computation is prohibitively expensive, we approximate $g_{\mathcal{C}}(s)$ by performing post-generation retrieval from the model's pre-training corpus, $\mathcal{C}$, and then testing whether $s$ can be grounded in the retrieved documents.
In practice, we compute the dense representation of $s$ and of each document in the corpus using the MiniLM-v2 \cite{wang2020minilm} sentence-Transformer \cite{reimers-gurevych-2019-sentence}. 
We perform an exact search for the top 5 pre-training documents $R_{\mathcal{C}}(s)$ closest to $s$ in terms of cosine similarity using the FAISS library \cite{johnson2019billion}. 
The grounding $g_{R_{\mathcal{C}}(s)}(s)$ of $s$ is checked against each retrieved document, as in Eq. \ref{eq:g}. We report additional experimental details in Appendix \ref{appendix:ptc_retrieval}.

\paragraph{Groundedness scores.} To quantify the groundedness of statements generated by the model, we calculate the fraction of sentences in the set $\mathcal{S}$ that are grounded in the retrieved or pre-training documents. Specifically, we compute the following expression:
\begin{align*}
 \frac{1}{|\mathcal{S}|} \big| \{s \in \mathcal{S} : \mathrm{condition}(s) \} \big|
\end{align*}
where $\mathrm{condition}$ denotes the grounding condition applied to each statement $s$. Based on this formulation, we compute groundedness scores for:
\begin{itemize}[leftmargin=15pt]
    \item the \textbf{retrieved documents only}, considering $s$ for which $g_{R(q)}(s) = 1 $ and $ g_{R_{\mathcal{C}}(s)}(s) = 0$,
    \item the \textbf{pre-training corpus only}, $s$ meeting the condition $g_{R(q)}(s) = 0$ and $g_{R_{\mathcal{C}}(s)}(s) = 1$,
    \item \textbf{both} ($g_{R(q)}(s) = 1$ and $g_{R_{\mathcal{C}}(s)}(s) = 1$),
    \item \textbf{none} ($g_{R(q)}(s) = 0$ and $g_{R_{\mathcal{C}}(s)}(s) = 0$).
\end{itemize}

\subsection{Experimental Setup}
\paragraph{Datasets.} 
We perform experiments on three datasets:
\begin{enumerate}
    \item \textbf{ASQA}~\cite{stelmakh-etal-2022-asqa} - ambiguous factual questions that have multiple correct answers. The desired model behavior includes providing a long-form generation that discusses all the correct answers.
    \item \textbf{HotpotQA}~\cite{yang-etal-2018-hotpotqa} - multi-hop question answering that requires reasoning over multiple entities in Wikipedia. The desired model behavior includes explicitly providing its reasoning in addition to the correct answer.
    \item \textbf{StrategyQA}~\cite{geva2021did} - multi-hop question answering, similar to HotpotQA. The correct answers to questions in this dataset are either True or False.
\end{enumerate}
On ASQA, for each question, we perform dense retrieval (using GTR; \citealp{ni-etal-2022-large}) from Wikipedia and include $k=3$ retrieved paragraphs in the model's prompt (example provided in Appendix \ref{appendix:prompts}). For HotpotQA and StrategyQA, instead of performing retrieval, we supply the model with the documents from which the correct answer can be determined 
(i.e., $R$ is an oracle function that retrieves necessary and sufficient information).
We do this to mitigate the effects of poor retrieval, since the reasoning over multiple documents required by these datasets makes correct retrieval necessary for reasonable performance.

\paragraph{Models.} We experiment with four different families of Transformer-based pre-trained language models: Pythia \cite{biderman2023pythia}, Falcon \cite{refinedweb}, MPT \cite{MosaicML2023Introducing}, and Silo \cite{min2023silo}. Post-generation retrieval is carried out on the whole training corpora for Pythia (the Pile; \citealp{gao2020pile}) and Silo (Open-license Corpus; \citealp{min2023silo}). For the MPT and Falcon models, we retrieve from the C4 dataset \cite{raffel2020exploring}, which represents $\sim$60\% of the training data used for MPT (in terms of \# of tokens), and was created in a similar way to the Falcon's training corpus \cite{refinedweb}.

\paragraph{Grounding.}Similar to prior work \cite{bohnet2022attributed, gao-etal-2023-enabling}, we assess groundedness using a natural language inference-based approach, which was shown to have a strong correlation with human judgment \cite{rashkin2023measuring, gao-etal-2023-enabling, chen2023understanding}. In particular, we use TRUE \cite{honovich-etal-2022-true-evaluating}, a T5-11B \cite{raffel2020exploring} model trained on a set of natural language inference datasets to automatically determine whether a generated statement is supported by a given text passage.
We carry out a manual validation of the TRUE model, in which we provide a small set of annotators with 100 instances of $(q, s, R(q), R_{\mathcal{C}}(s), g_{R(q)}(s), g_{R_{\mathcal{C}}(s)}(s))$. The annotators are asked to judge whether the grounding model's predictions $g_{R(q)}(s)$ and $g_{R_{\mathcal{C}}(s)}(s)$ are correct (i.e., whether the generated statement is correctly determined to be grounded or ungrounded with respect to the considered sources). We observe that the annotators agree with the model in 82\% of the cases overall and in 98\% of the correct but ungrounded cases. We provide additional details about the groundedness verification procedure and its manual validation in Appendix \ref{appendix:groundedness-verification}.

\section{How Frequently are Generations Grounded?}
\begin{figure}[t]
    \centering
    \includegraphics[width=\columnwidth]{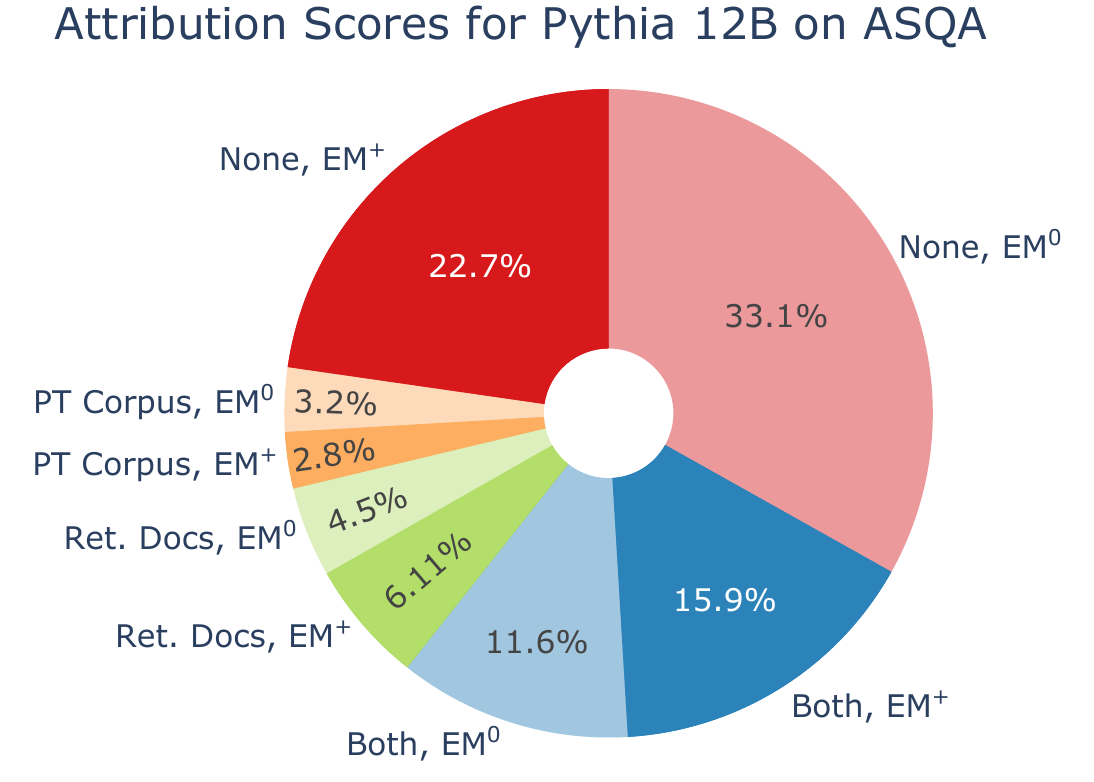}    \caption{\textbf{Groundedness \& correctness.} Each of the 8 sectors in the chart corresponds to a specific combination of groundedness (in the retrieved documents, pre-training data, both, or neither) and EM correctness (either belonging to $\mathrm{EM}^0$ or $\mathrm{EM}^+$). The area of a sector corresponds to the fraction of all model-generated sentences over all ASQA test examples that exhibit that groundedness-correctness combination.}
    \label{fig:pie}
\end{figure}

We begin our analysis by measuring the rate at which models of various sizes and families generate sentences that are grounded in the retrieved documents as well as the pre-training data.
As an example, consider Figure \ref{fig:pie}, which provides a visualization of the sentences generated by the Pythia 12B model on the ASQA dataset divided into 8 sectors. 
Each sector represents one group in the cross-product of the following categories:
\begin{itemize}
    \item whether a sentence could be grounded in the retrieved documents, the pre-training data, both, or neither;
    \item and whether the sentence was part of a long-form generation that was deemed incorrect ($\mathrm{EM}^0$), or not ($\mathrm{EM}^+$), according to exact match.
\end{itemize}
We expect that most sentences that are part of $\mathrm{EM}^+$ will be grounded in either the retrieved documents or the pre-training corpus (or both), since those documents represent the source of the model's correctness.
For sentences that are part of $\mathrm{EM}^0$, we make no assumptions on the frequency of grounding.
That is because incorrect answers could arise from the model emitting ungrounded sentences, or grounded sentences that are off-topic or otherwise incorrect.

Figure \ref{fig:pie} reveals that overall, for Pythia 12B, $\sim$44\% of the sentences generated are grounded in at least one of the two sources considered (i.e., blue, green, and orange sectors), and that $\sim$48\% belong to $\mathrm{EM}^+$ (i.e., the darker-color sectors).
Interestingly, we observe that nearly half of the generated sentences belonging to $\mathrm{EM}^+$ cannot be grounded in the retrieved documents or pre-training corpus. This is unexpected since, by virtue of containing at least some of the ground-truth answers, the model demonstrates that it may have access to relevant information. Upon inspection, we find that the high proportion of ungrounded sentences in these answers contain ground-truth named entities or text snippets that are presented in a nonsensical or factually incorrect manner.
We provide examples of such outputs in Section \ref{sec:examples}. We note that sentences that cannot be grounded via our methods could be the result of sub-optimal retrieval in the pre-training corpus, or errors from the TRUE (grounding) model. We elaborate more on these concerns in Appendices \ref{appendix:ptc_retrieval} and \ref{appendix:groundedness-verification}.

Additionally, Figure \ref{fig:pie} also shows that roughly one-fourth of the generations can be grounded in both the pre-training and retrieved document. This is likely due to overlap between the pre-training and retrieval corpora, or the appearance of common knowledge present in both corpora.

\begin{figure}[t]
    \centering
    \includegraphics[width=\columnwidth]{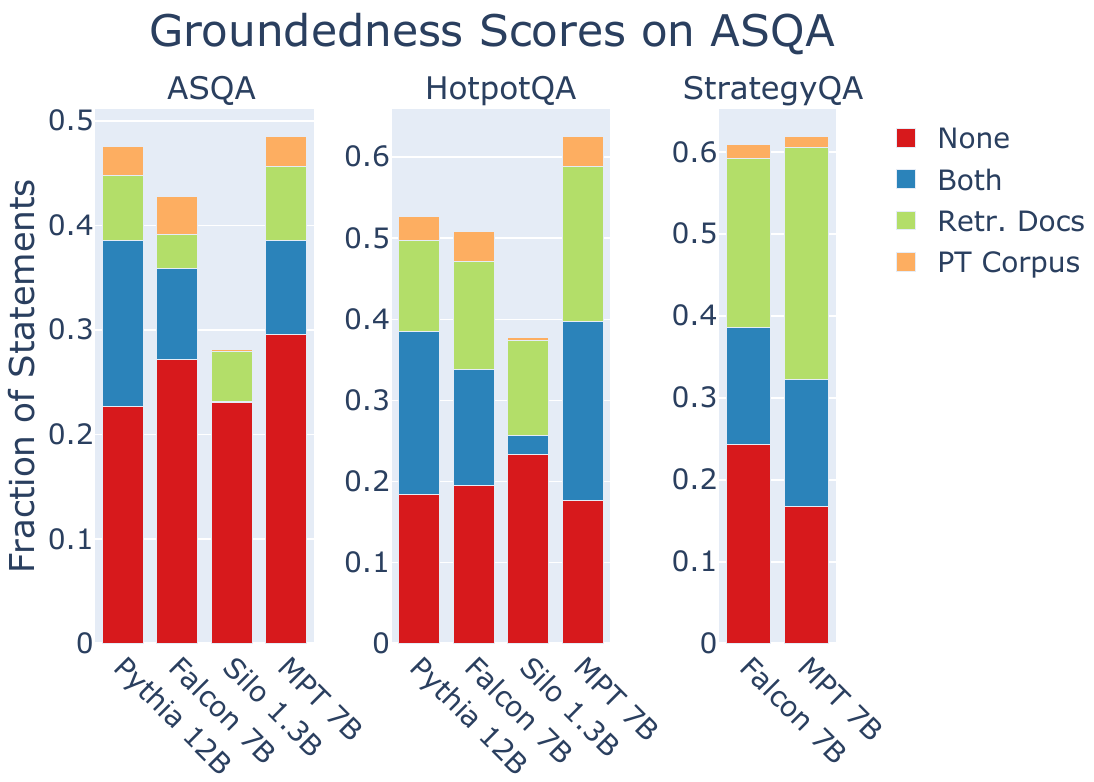}
    
    \caption{\textbf{Groundedness across datasets.} The height of each bar represents the fraction of generated sentences that belong to partially correct generations. A significant fraction of these sentences are not grounded in either the  retrieved or pre-training documents.}
    \label{fig:all-models}
\end{figure}

\paragraph{Does ungrounded content appear consistently?}
For the remainder of this analysis, %
we focus on outputs that contain ground-truth answers (i.e., $\mathrm{EM}^+$), as they might represent subtler and more interesting cases of undetected hallucination.
Figure \ref{fig:all-models} visualizes of frequency of grounded sentences in correct and partially correct answers on ASQA, HotPotQA, and StrategyQA for four models from the four families considered: MPT 7B, Falcon 7B, Silo 1.3B, and Pythia 12B. 
Note that the height of each bar represents the fraction of generated sentences that belong to long-form generations containing at least 1 ground-truth answer.

On all three datasets and all models considered, we observe that a substantial fraction of the outputs in $\mathrm{EM}^+$ are not grounded in the retrieved documents or in the pre-training data.\footnote{Results for Pythia and Silo are omitted for StrategyQA due to their accuracy being marginally better than a random chance, precluding meaningful analysis.} This consistency across different models and datasets indicates a prevalent pattern where models are able to generate correct answers that are found in sentences that are not directly supported by the retrieved documents or pre-training data.

\section{What Factors Affect Groundedness?}
In this section, we study the interplay between the tendency of models to generate grounded content and three factors: the size of the model (Section \ref{sec:model-size}), the decoding strategy (Section \ref{sec:decoding-strategy}), and instruction-tuning (Section \ref{sec:instruction-tuning}).

\subsection{Model Size}
\label{sec:model-size}
\begin{figure}[t]
    \centering
    \includegraphics[width=\columnwidth]{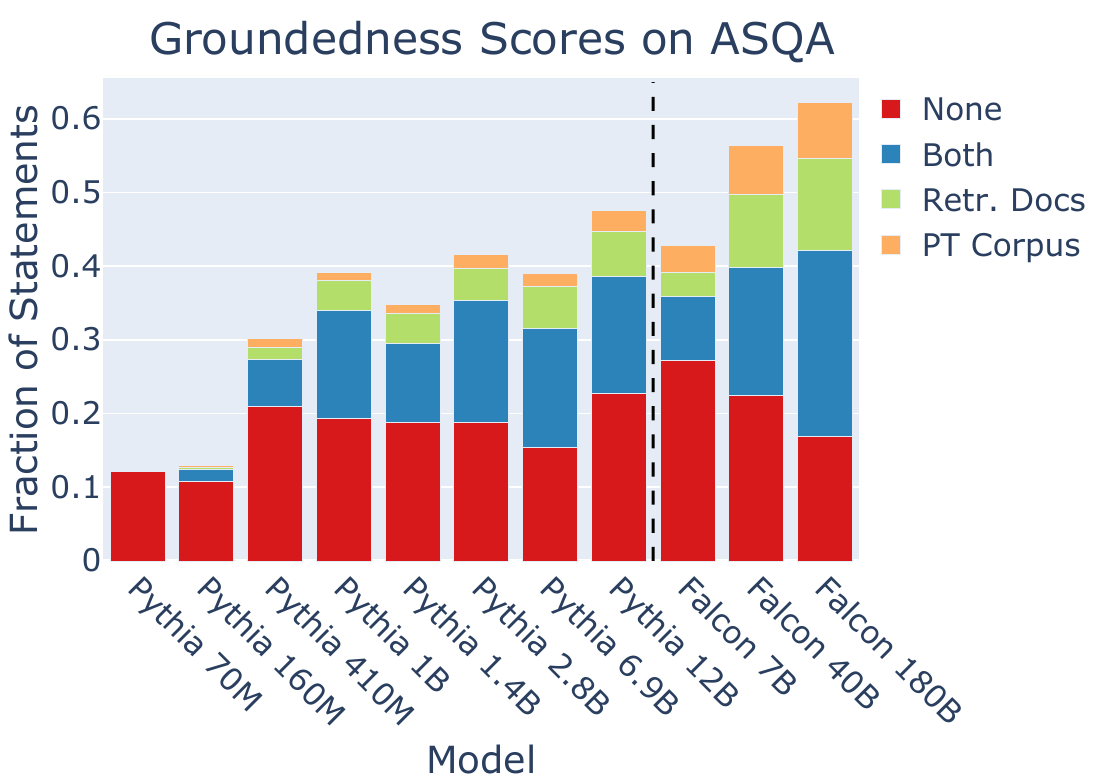}
    \caption{\textbf{Groundedness by size.} As before, the height of each bar represents the fraction of generated sentences that belong to partially correct generations. Increased model size correlates with an increase in the number of sentences in $\mathrm{EM}^+$, but also an increase in groundedness.}
    \label{fig:size}
\end{figure}

For smaller models, particularly Pythia 70M and Pythia 410M, the majority of generated sentences cannot be grounded in either the retrieved documents or the pre-training corpus (Figure \ref{fig:size}).
Interestingly, while they do not have a clear grounding to either the documents in the context or the pre-training corpus, the responses generated by these models occasionally match tokens from the ground truth answers.
A possible interpretation of this result is that these models may rely more on internal heuristics or pattern-matching capabilities rather than effectively using external information or learned knowledge \cite{elazar2022measuring, mccoy-etal-2019-right}.

Pythia models in the range 1-12B generate an increased fraction of grounded output compared to their smaller counterparts. However, no stark trend is observed within this size range.
Conversely, for significantly larger models (Falcon 40B and 180B), there is a clear increase in the proportion of content that can be grounded to the provided context or the pre-training corpus. This indicates that larger models are better equipped to integrate and utilize external information from the provided context and their extensive pre-training.
However, it is important to note that even with the largest models, there remains a non-negligible fraction of generated sentences that are part of $\mathrm{EM}^+$ but cannot be grounded in either the retrieved documents or the pre-training corpus.\looseness=-1

\subsection{Decoding Strategy}
\label{sec:decoding-strategy}

\begin{figure}[t]
    \centering
    \includegraphics[width=\columnwidth]{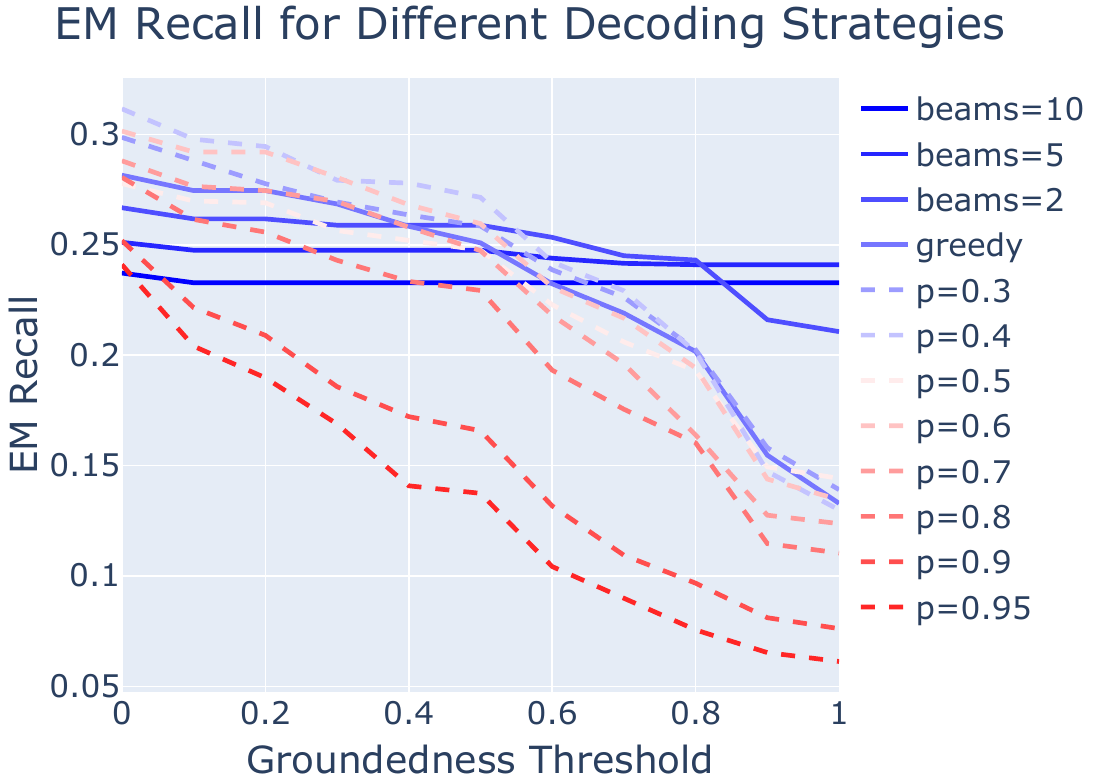}
    \caption{\textbf{Groundedness by decoding.} Exact Match (EM) scores against the minimum fraction of sentences required for a model generation to be considered valid (\textit{groundedness threshold}). As the grounding threshold tightens, the EM scores for random sampling quickly degrade. The scores for beam search, however, remain roughly unaltered, indicating a higher level of groundedness. Results obtained with Pythia 12B on ASQA.}
    \label{fig:decoding}
    \vspace{-1mm}
\end{figure}

We measure the impact of the decoding algorithm on the frequency of grounded sentences as well as correctness.
In particular, we test readily available decoding strategies: greedy decoding, nucleus sampling \cite{holtzman2019curious}, and beam search.
For nucleus sampling we vary the \texttt{top\_p} parameter; for beam search, we test beam widths 2, 5, and 10.

Figure \ref{fig:decoding} illustrates the impact of employing various decoding methods with the Pythia 12B model on the ASQA dataset.
In the Figure, the $x$-axis represents a \emph{groundedness threshold}, i.e., the minimum fraction of sentences in a model generation that must be grounded (in either retrieved or pre-training documents) for that generation to be considered valid. That is, at $x=1.0$, all sentences in a generation must be grounded for the generation to be valid. For any groundedness threshold $x$, the corresponding $y$-value represents the average Exact Match (EM) score across generations, where invalid generations automatically get a score of 0. 

Intuitively, as the groundedness threshold becomes more stringent (i.e., as we require a higher fraction of sentences to be grounded), the EM scores should decrease.
Indeed, this trend is observed for greedy decoding and nucleus sampling. 
However, an interesting deviation from this trend is observed in the case of beam search decoding. Unlike the other strategies, the EM scores for beam search do not exhibit a significant decline as the groundedness threshold increases.
In particular, the decline is less steep as the number of beams increases.
This result shows that while beam search may initially show lower EM scores compared to nucleus sampling without considering groundedness, its effectiveness emerges when groundedness is taken into account.
A possible explanation for this phenomenon can be identified in the tendency of beam search to give a higher likelihood to sequences that previously appeared in the model input \cite{holtzman2019curious}. Since in our setting the retrieved documents are provided as an input sequence to the model, greedy and nucleus sampling might assign a higher probability to grounded sentences.

\begin{figure}[t]
    \centering
    \includegraphics[width=\columnwidth]{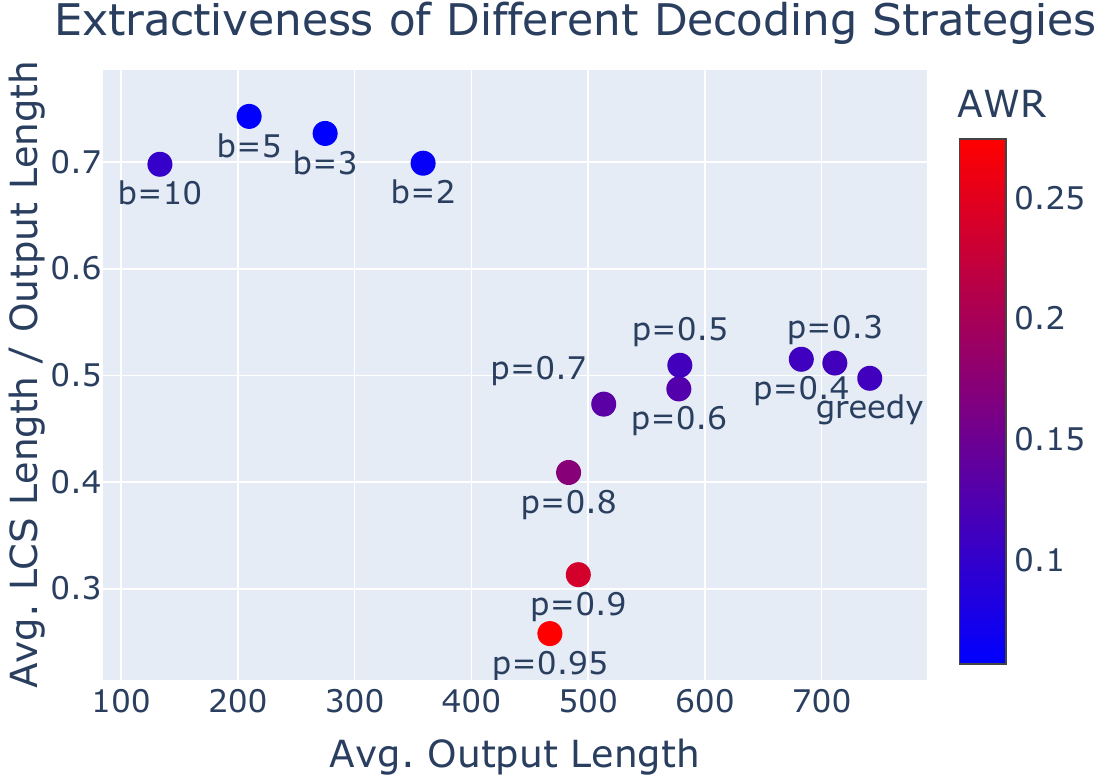}
    \caption{\textbf{Extractiveness by decoding.} Average ratio between the length of the LCS and the length of the model output ($y$-axis) against the average length of the output ($x$-axis). Length is measured as the number of characters. The color scale illustrates the average ratio of abstracted words (AWR) for the different decoding strategies: beam search (b), greedy, and nucleus sampling (p). Results obtained on ASQA with Pythia 12B.}
    \label{fig:extractiveness}
\end{figure}

To clarify this aspect, we carry out analyses of the models’ \textit{extractiveness} (i.e., the tendency of a model to replicate portions of text verbatim from the retrieved documents). The results are reported in Figure \ref{fig:extractiveness}. By measuring the longest common substring (LCS) between the model outputs and the retrieved documents, and comparing this to the overall length of the model outputs, we find that a larger number of beams generally result in shorter outputs, but with proportionately longer common substrings. Additionally, we compute the proportion of abstracted words (i.e., words that do not appear in any of the documents included in the prompt) present in the model output and notice that it decreases as the decoding becomes less random (as the parameter $p$ in nucleus sampling decreases, as one would expect), but also that it becomes substantially smaller with beam search. These findings clarify how beam search affects groundedness and suggest a trade-off between extractiveness and groundedness in retrieval-augmented generation.

\subsection{Instruction Tuning}
\label{sec:instruction-tuning}

\begin{figure}[t]
    \centering
    \includegraphics[width=\columnwidth]{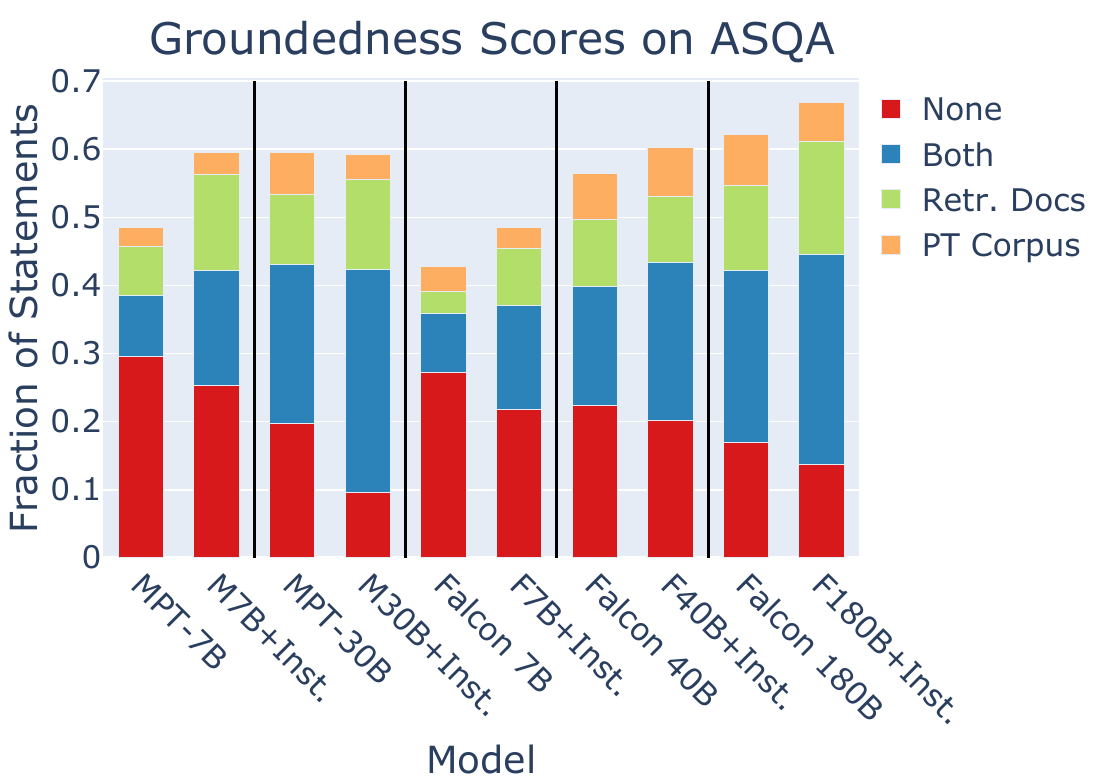}
    \caption{\textbf{Effect of instruction-tuning.} As before, the height of each bar represents the fraction of generated sentences that belong to $\mathrm{EM}^+$. Compared to the corresponding base models, instruction-tuned models tend to exhibit greater correctness as well as a larger fraction of grounded sentences.}%
    \label{fig:instruction}
\end{figure}

Figure \ref{fig:instruction} illustrates the impact of instruction tuning on the groundedness and correctness of sentences generated by various models on the ASQA dataset.
For instruction-tuned models, we observe a marked improvement in both the overall correctness and the fraction of grounded sentences. 
This trend holds across various model families and sizes,
suggesting that instruction tuning enhances the model's ability to generate content that not only contains ground-truth answers but is also more grounded in the provided context or pre-training data.

\begin{table*}[t]
\centering
\small
\begin{tabularx}{\textwidth}{X X X X c c}
    \toprule
     Query $q$ & Sentence $s$ & $d \in R(q)$ & $d \in R_{\mathcal{C}}(s)$ & $g_{R(q)}(s)$ & $g_{R_{\mathcal{C}}(s)}(s)$ \\
     \midrule 
      \multirow{7}{\hsize}{What bodies of water surround Cabo San Lucas?} & \multirow{7}{\hsize}{\dots The \textbf{Sea of Cortez} is part of the \textbf{Pacific Ocean}, which is connected to the Atlantic Ocean via the \textbf{Gulf of California}.} & \multirow{7}{\hsize}{ \dots El Arco, a natural stone arch over ocean in Cabo San Lucas where the Pacific Ocean and Gulf of California meet \dots} & \dots compared to the west side of the Baja Peninsula, the Sea of Cortez has warmer weather and a climate more in line with most people’s\dots & \multirow{7}{*}{ 0} &  \multirow{7}{*}{0} \\
    \midrule
    \multirow{7}{\hsize}{When did the Brisbane Broncos last win a premiership?} & \multirow{7}{\hsize}{The Brisbane Broncos last won a premiership in \textbf{2006}.} & \multirow{7}{\hsize}{\dots having endured twelve completed seasons without winning the premiership since their most recent success in 2006.} & \dots The Broncos did not start the 2006 season well, suffering a 36–4 defeat by the previous year's runners-up, the North Queensland\dots & \multirow{7}{*}{1} & \multirow{7}{*}{0} \\
    \midrule
    \multirow{7}{\hsize}{When will they finish building the Sagrada Familia?} & \multirow{7}{\hsize}{\dots The construction of this building began in 1882 and it will be finished in \textbf{2026}.} & \dots is large unfinished Roman Catholic church in Barcelona, designed by Catalan architect Antoni Gaudí (1852–1926). Gaudí's \dots  & \multirow{7}{\hsize}{\dots 's still being worked into today a goal of completion in 2026 a century after Gaudi's death\dots } & \multirow{7}{*}{0} & \multirow{7}{*}{1} \\
    \midrule
    \multirow{5}{\hsize}{Names of the rivers in the garden of Eden?} & \multirow{5}{\hsize}{The rivers in Eden were the \textbf{Pishon}, the \textbf{Gihon}, and the \textbf{Tigris}, and \textbf{Euphrates}.} & \dots divided into four branches: the Pishon, where the gold is; the Gihon; the Tigris; and the Euphrates. \dots & \dots leaving the garden branching out into four streams, the Euphrates, Gihon, Pison, and the Tigris. & \multirow{5}{*}{1} & \multirow{5}{*}{1} \\     
     \bottomrule
\end{tabularx}
    \caption{\textbf{Examples of generations} by Pythia 12B on ASQA, along with relevant portions of the retrieved documents and the predictions from the groundedness-verification model ($g_{R(q)}(s)$ and $g_{R_{\mathcal{C}}(s)}(s)$). The sentences were selected among the generations containing one or more ground-truth answers (reported in bold), i.e., they belong to $\mathrm{EM}^+$.}%
    \label{table:examples}
\end{table*}

\begin{table}[t]
\centering
\small
\begin{tabularx}{\columnwidth}{X X}
    \toprule
     \textbf{Query} $q$ & \textbf{Generated Sentence} $s$  \\
     \midrule
      \multirow{8}{\hsize}{How many times did Brazil win the FIFA World Cup?} & Brazil has won the World Cup 5 times. They won in 1958, 1962, 1970, 1994, and 2002. \textcolor{Red}{\textbf{The only other country that has won as many times as Brazil is Germany}} (also known as West Germany). \\
      \midrule
      \multirow{8}{\hsize}{Who did the artwork for Pink Floyd's wall?} & The artwork for Pink Floyd's Wall was created by Gerald Scarfe. The artwork on the cover of The Wall features a uniform row of white bricks \textcolor{Red}{\textbf{with a hammer in the foreground}}.  \\
      \midrule
       \multirow{7}{\hsize}{The director of the romantic comedy ``Big Stone Gap'' is based in what New York city?} & Adriana Trigiani is the writer of the romantic comedy Big Stone Gap. \textcolor{Red}{\textbf{She was born and raised in New York City}}. She works in Greenwich Village, New York City.\\
     \bottomrule
\end{tabularx}
    \caption{\textbf{Examples of generations} by Falcon 180B on ASQA (first two rows) and HotpotQA (last row). The generated content features a mixture of correct and fabricated content (the latter highlighted in \textcolor{Red}{\textbf{red}}): Germany won the World Cup 4 times, not 5, \textit{The Wall}'s album cover does not feature a hammer in the foreground, and Adriana Trigiani was actually born in Virginia.}
    \label{table:falcon_examples}
\end{table}

\section{Examples of Model Output}
\label{sec:examples}
In Table \ref{table:examples}, we present a handful of representative sentences generated by the Pythia 12B model on the ASQA dataset. 
Additionally, we supply relevant portions of the retrieved documents, $R(q)$, (from Wikipedia) as well as the documents retrieved from the pre-training corpus, after the model generates an answer.
Overall, the selected sentences demonstrate the model's ability to generate content that, while sometimes correctly includes ground-truth answers, showcases varying degrees of groundedness. 

Consider the example in the first row of Table~\ref{table:examples}.
In this instance, the model includes the correct named entities within its generated text.
However, it also produces an ungrounded (and incorrect) claim that the Pacific Ocean is ``\textit{connected to the Atlantic Ocean via the Gulf of California.}'' 
This statement is factually incorrect and represents a clear case of the model generating a plausible-sounding but erroneous connection between entities. 

Inspecting the model-generated answers, we observed a common trend in which models blend fabricated content with factually correct statements.
This trend is still present in larger models, whose generations are overall more grounded than small models but might feature a melding of fact and hallucination that can be more subtle.
In Table \ref{table:falcon_examples}, we provide some examples obtained with Falcon 180B.
These instances illustrate the models' ability to accurately retrieve and use specific terms from the training data or the retrieved documents while also highlighting the challenge of ensuring that the relationships and contexts it generates are factual.

\section{Related Work}

\paragraph{Long-form Question Answering.}
LFQA (also referred to as generative QA)  \cite{fan-etal-2019-eli5} is a question-answering task in which the goal is to generate---rather than extract---the correct answer to a question, usually by drawing from provided passages.
The majority of the work in LFQA involves a human evaluation process aimed at measuring the factual correctness of generated answers \cite{su-etal-2022-read, nakano2021webgpt, krishna-etal-2021-hurdles}.
In the absence of human evaluation, the quality of an answer is typically evaluated using automatic metrics such as ROUGE \cite{lin-2004-rouge}. However, these metrics require human-annotated answers, and, more importantly, they fail to pick up hallucinated content \cite{xu-etal-2023-critical, ji2023survey, krishna-etal-2021-hurdles}. %
Other approaches, such as factual consistency \cite{durmus-etal-2020-feqa, wang-etal-2020-asking}, are used to measure the faithfulness of a generated summary to a corresponding set of source documents.
Similar strategies were subsequently applied to evaluate dialogue systems \cite{honovich-etal-2021-q2} and to edit and improve the output of retrieval-augmented language models \cite{gao-etal-2023-rarr}.

\paragraph{Groundedness \& Attribution.}
In light of the importance of mitigating hallucination, work on evaluating groundedness has enjoyed significant attention \cite{li2023survey}.
Initial work develops the \emph{attributable to identified sources} (AIS) score, which represents a human evaluation quantifying the degree to which generated text adheres to its cited sources~\cite{rashkin2023measuring}. 
Later work demonstrates that AIS can be well-approximated using a model trained for predicting entailment, as in natural language inference~\cite{honovich-etal-2022-true-evaluating}.
However, in these studies, groundedness is often computed against provided passages~\cite{bohnet2022attributed, gao-etal-2023-enabling, yue2023automatic}. 
In contrast to previous studies that treat the accuracy of correct entities and output groundedness as distinct aspects \cite{gao-etal-2023-enabling}, our research delves into exploring their interconnection.

More similar to our work are those that measure groundedness against documents retrieved by a web search API~\cite{liu2023evaluating, gao-etal-2023-rarr, chen2023understanding}.
In some cases, such as checking the existence of a generated reference, this is an appropriate strategy~\cite{agrawal2023language}.
But in the general case, we argue that such an approach can be problematic because of the varying quality, factuality, and relevance of internet search results.\looseness=-1

Another line of work explores the relationship between a model's generated text and its pre-training data.
For example, one study measures how often models repeat content verbatim from their pre-training corpora \cite{mccoy2023much}.
Similarly, other works study the provenance of the model-generated content within pre-training corpora but rely on gradient-based methods \cite{han2022orca} or metrics based on n-gram overlap \cite{weller-etal-2024-according}.
Others analyze model generations with the intent to characterize the extent to which they can be attributed to a model's parametric memory vs. additional information provided at inference.
However, this is measured by either constructing prompts that contain sentences that conflict with information in the pre-training data~\cite{longpre-etal-2021-entity}, or by drawing on correlations with respect to the rarity of the entities produced in model generations~\cite{mallen-etal-2023-trust}.
Unlike these works, we verify whether the model output can be supported by passages retrieved from the pre-training corpus, as well as the context supplied during inference.

\section{Conclusion}
This study analyzes the rate at which the long-form output produced by retrieval-augmented LLMs is grounded in retrieved documents and pre-training data.
Through empirical analysis across various models and datasets, we highlight the propensity of LLMs to blend correct information with hallucinated content. Our findings indicate that this tendency is prevalent across different model sizes and persists even in the largest models available.
Our analyses reveal that while larger models generally produce more grounded content, they are not immune to generating ungrounded information. We observed that instruction tuning and beam search decoding reduce ungrounded sentence generation. Aligned with the results of concurrent research \cite{choi-etal-2023-kcts}, our findings point to specialized decoding algorithms being good candidates for significantly reducing hallucination.

\section*{Limitations}
We identify a handful of limitations of our work below.

\paragraph{Imperfect Retrieval from Pre-training Corpus.} Given the size of pre-training corpora, it is possible for our approach to exhibit false negatives. 
That is, when attempting to retrieve passages in the pre-training corpus that ground a model-generated sentence, we may incorrectly conclude that the generated sentence is not grounded in the pre-training corpus. 
This is a result of retrieval being imperfect.
Despite this, we suspect that the rate of these false negatives is low given: a) manual inspection of the ungrounded sentences and b) the relatively high true positive rate, i.e., that rate at which we successfully ground generated sentences in pre-training documents.
\paragraph{Scattered Correct Information.} When attempting to ground a model-generated sentence, our approach considers each source document independently. However, this is limited when a generated sentence amalgamates information from multiple sources, in a way no single source fully supports. An enhanced method, potentially examining the concatenation of multiple source documents, could address this issue, and we propose this as an area for future research. A related---and more general---limitation is our reliance on a grounding model, which is also imperfect. 
\paragraph{Dependence on Pre-training Data Availability.} Our methodology relies on accessing the pre-training corpus or a dataset containing most of the documents contained therein. This dependence is a significant limitation, especially for models where the pre-training data is not readily available or is incomplete. Indeed, this requirement significantly limits the family of models we use in our experiments.

\section*{Acknowledgments}
Alessandro would like to thank Ari Kobren for providing an incredibly generous amount of feedback, guidance, and practical support throughout the project. He would also like to thank the whole Machine Learning Research Group at Oracle Labs, most notably Adam Pocock, Qinlan Shen, and Michael Wick for providing feedback and helping with manual annotation, Jason Peck and Jeffrey Alexander for the support with the computing infrastructure, and Stephen Green for practical support.
Alessandro is supported by armasuisse Science and Technology through a CYD Doctoral Fellowship. 

\bibliography{bib/anthology,bib/RALMs}
\bibliographystyle{acl_natbib}

\clearpage

\appendix

\section{Prompting Details}
\label{appendix:prompts}
To elicit models to generate long-form answers leveraging the retrieved documents, we construct a few-shot prompt with exemplars of question-documents-answer triples. In particular, we use 3-shot prompts on HotpotQA and StrategyQA and 2-shot prompts on ASQA (due to the longer documents used for this dataset). In Table \ref{table:prompts}, we provide an example of a prompt used on ASQA.

\section{Pre-training Corpus Retrieval}
\label{appendix:ptc_retrieval}

\subsection{Retrieval Details}
The retrieval procedure from the pre-training corpora was carried out by first dividing the corpus into passages of 768 contiguous characters. Then, each passage was embedded using the MiniLM-v2 \cite{wang2020minilm} sentence-Transformer \cite{reimers-gurevych-2019-sentence}.  This procedure was carried out in parallel by 12 64-CPU computing nodes in parallel and took $\sim$24 hours for each corpus. Finally, given a sentence generated by a model, a search for the 5 most relevant passages in the corpus was performed using the FAISS library \cite{johnson2019billion}. The search was carried out in parallel by 20 computing nodes on different subsets of the corpus and took $\sim$12 hours for all sentences generated by a model on a dataset.

\subsection{Validation of the Retrieval Procedure}
Given the size of pre-training corpora, it is possible for our retrieval approach to produce false negatives. However, we believe that the rate of these false negatives is low for two main reasons. First, the rate of positives (i.e., the rate at which we successfully ground generated sentences in pre-training documents) is relatively high (e.g., 34\% and 48\% of the overall statements for Pythia 12B and Falcon 180B, respectively). Second, when manually inspecting ungrounded sentences, we notice that a large number of them are nonsensical or contain fabricated information.

Moreover, we carry out an additional validation study. Recognizing the impracticality of conducting a comprehensive search across the entire pre-training corpus to definitively show the absence of supporting text for a given claim, we opt for a focused approach. We analyze a random subset of the generated statements that are judged ungrounded and manually determine whether each sentence is factually correct. We found that 40 out of 50 instances inspected contain factually incorrect information. We report some examples of generations by Pythia 12B in Table \ref{table:factual_mistakes}.

While the factual incorrectness of a sentence does not definitively rule out support from pre-training documents, it strongly suggests their absence. We therefore believe that our method of retrieving from the pre-training corpus represents a reasonable approximation for verifying groundedness in this context.

\begin{table*}[t]
\centering
\small
\begin{tabularx}{\textwidth}{X X X}
    \toprule
     \textbf{Query} $q$ & \textbf{Generated Sentence} $s$ & \textbf{Comment} \\
     \midrule
      \multirow{5}{\hsize}{Who's the highest paid NBA player in 2017?} & Kobe Bryant became the third player to achieve the milestone of earning \$40-Million in the 2016–17 season, surpassing the record of 29-Million he had earned the previous season. & \multirow{5}{\hsize}{Kobe Bryant retired in 2016.}\\
      \midrule
      \multirow{4}{\hsize}{How many medals did Australia win in the 2000 Olympics?} & The host nation had never entered in the Summer Olympics before, the Sydney Olympics being Australia's first appearance. & \multirow{4}{\hsize}{Australia's first appearance in the Olympics was in 1896.} \\
      \midrule
       \multirow{8}{\hsize}{What is the oldest company in the Dow Jones index?} & In the early 2000s, the Dow Jones Industrial Average was composed of the following nine companies: The components of the DJIA have changed 52 times since its beginning on May 26, 1896. General Electric had the longest continuous presence on the index, beginning in 1907 and ending in 2018. & \multirow{8}{\hsize}{The last sentence is correct and grounded in a retrieved document, but the statement ``In the early 2000s, the Dow Jones Industrial Average was composed of the following nine companies'' is hallucinated.} \\
     \bottomrule
\end{tabularx}
    \caption{\textbf{Examples of ungrounded generations.} The generated content features hallucinated details that render the statements factually incorrect. The examples are obtained with Pythia 12B on ASQA.}
    \label{table:factual_mistakes}
\end{table*}

\section{Groundedness Verification Method}
\label{appendix:groundedness-verification}
We performed inference with the TRUE model \cite{honovich-etal-2022-true-evaluating} following previous work that employed the model for attribution verification \cite{gao-etal-2023-rarr, bohnet2022attributed, gao-etal-2023-enabling, chen2023understanding}: the model was prompted with a concatenation of a potential supporting document (preceded by the string ``\texttt{premise:}'') and an LLM-generated statement (preceded by the string \texttt{“hypothesis:”}). If the model’s output is ``\texttt{1}'', then the generated statement is considered grounded in the supporting document, otherwise not.

\subsection{Evaluation of the Verification Method}
\label{appendix:manual-evaluation}
The validation of the model was carried out by a team of 5 annotators (consisting of the author and collaborators), each of whom was assigned a set of 20 instances of $(q, s, R(q), R_{\mathcal{C}}(s), g_{R(q)}(s), g_{R_{\mathcal{C}}(s)}(s))$, where
\begin{itemize}
    \item $q$ is a question,
    \item $s$ is a statement belonging to the model-generated answer to $q$,
    \item $R(q)$ are the documents retrieved to augment the LLM generation,
    \item $R_{\mathcal{C}}(s)$ are the pre-training documents retrieved post-generation,
    \item $ g_{R(q)}(s) $ and $ g_{R_{\mathcal{C}}(s)}(s)$ are the groundedness predictions of the TRUE model with respect to each of the potential supporting document.
\end{itemize}
The 20 instances are sampled at random, making sure that an equal amount of instances comes from each of the categories: ungrounded, grounded in the pre-training documents only, grounded in the pre-generation retrieved documents only, and grounded in both types of documents. For each instance, an annotator determines whether the predictions of the groundedness model are correct or not. A score of 1 was assigned by the annotator if the groundedness model accurately identified the supporting document subset (which could be none, in the case of ungrounded content) for the given LLM-generated statement. Conversely, a score of 0 was given if the model failed to correctly identify the supporting documents.

\section{Additional Experimental Details}
\label{appendix:exp_details}

\paragraph{Computing infrastructure.}
All experiments with models in the size range 70M-12B were carried out using a single 40GB Nvidia A100. Generations with MPT 30B and Falcon 40B were obtained using four 40GB Nvidia A100s, and with Falcon 180B using eight 40GB Nvidia A100s. The runtime for each model on each dataset was $\leq$12 hours.

\paragraph{Licenses.} For our analyses we three QA datasets (ASQA, HotpotQA, and StrategyQA) and three pre-training corpora (the Pile, C4, and OLC). ASQA is available under the Apache 2.0 license, StrategyQA, the Pile, and OLC are available under the MIT license, HotpotQA is available under CC-BY,  and C4 is released under the terms of ODC-BY.

\section{Additional Results}
\label{appendix:results}
\begin{figure}[t]
    \centering
\includegraphics[width=\columnwidth]{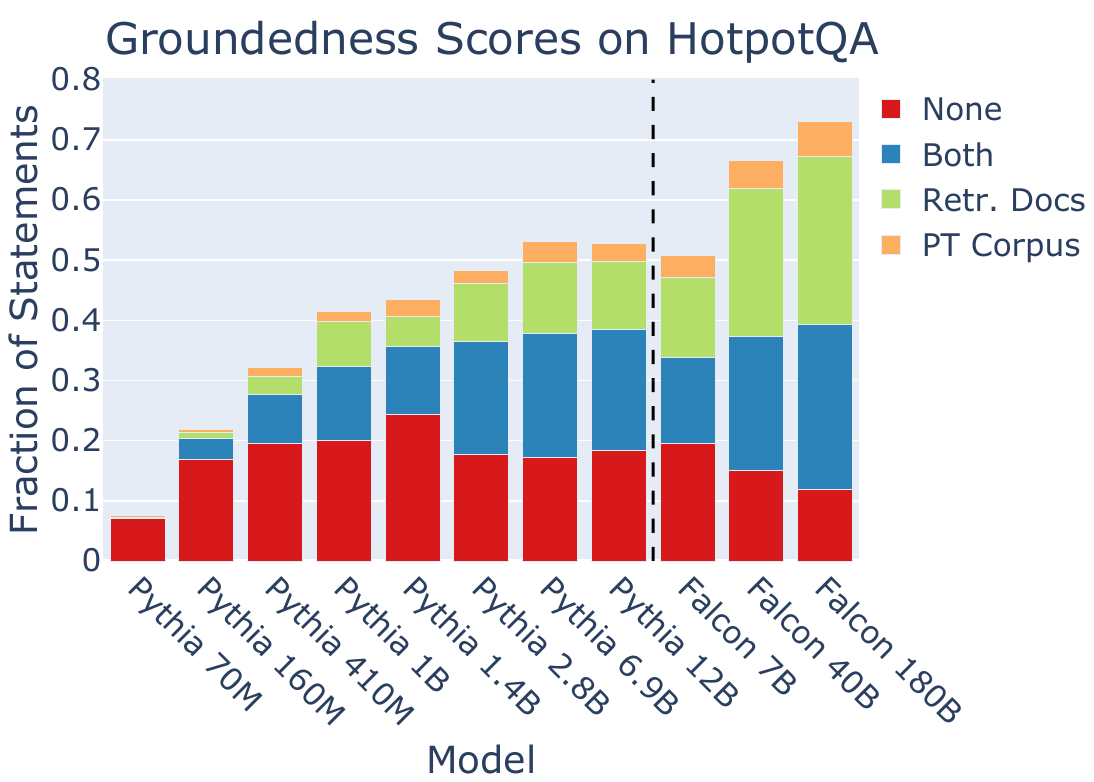}
    \caption{\textbf{Groundedness by size.} The results are consistent with the one obtained on ASQA.}
    \label{fig:size-hotpotqa}
\end{figure}

\begin{figure}[t]
    \centering
    \includegraphics[width=\columnwidth]{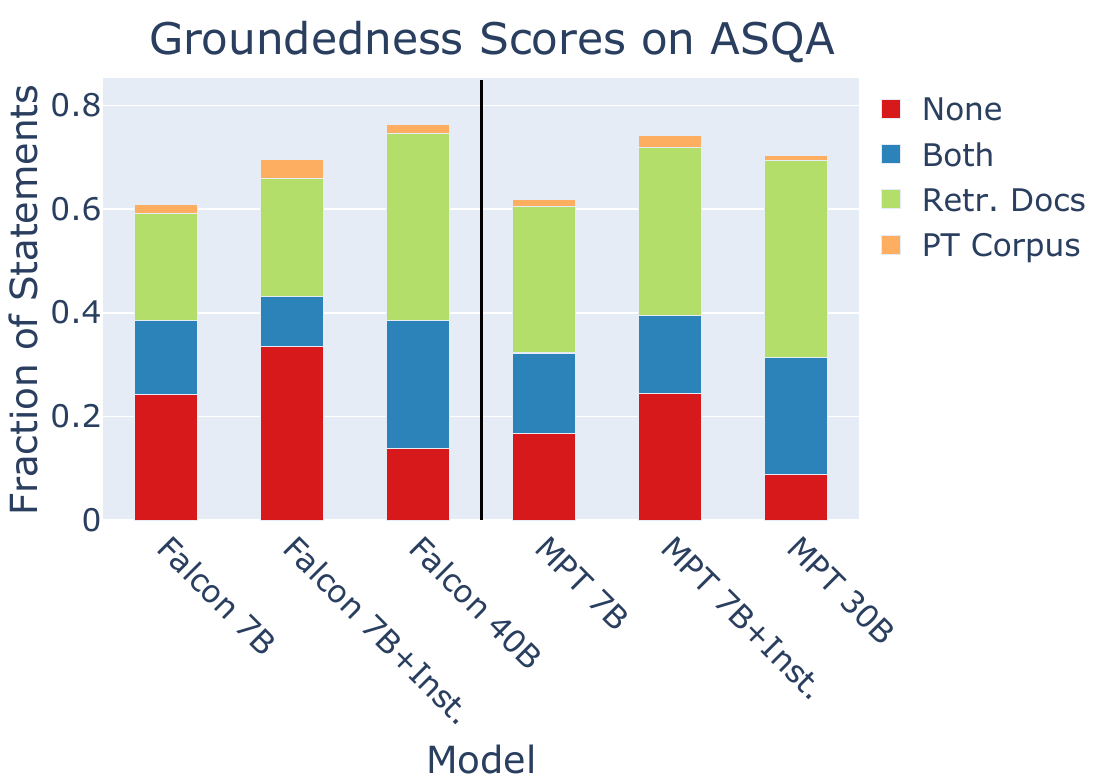}
    \caption{\textbf{Groundedness on StrategyQA.} As in previous figures, the size of each bar represents the fraction of generated sentences that belong to partially correct generations.}
    \label{fig:strategyqa}
\end{figure}

In Figure \ref{fig:size-hotpotqa}, we report the groundedness scores computed for the Pythia and Falcon models with different sizes on HotpotQA. We observe similar trends to the ASQA setting. 
Figure \ref{fig:strategyqa} illustrates the groundedness scores obtained with different Falcon and MPT models on StrategyQA.

\begin{table*}[t]\small
\resizebox{\textwidth}{!}{
    \centering
    \begin{tabularx}{\textwidth}{ X }
    \toprule[0.1em] \\
Instruction: Write an accurate, engaging, and concise answer for the given question, possibly using the provided search results (some of which might be irrelevant).\\
\\
Question: Who played galen in planet of the apes?\\
\\
Document [1](Title: Planet of the Apes): installment. Jacobs died on June 27, 1973, bringing an end to the APJAC Productions era of the "Planet of the Apes" franchise. Former Fox executive Stan Hough took over as producer for the television project, titled "Planet of the Apes". CBS picked up the series for its 1974 autumn lineup. Ron Harper and James Naughton played Alan Virdon and Peter Burke, two 20th-century American astronauts who pass through a time warp to a future where apes subjugate humans (unlike the original film, the humans can speak). Roddy McDowall returned to the franchise as Galen, a chimpanzee who joins the astronauts.\\
Document [2](Title: Planet of the Apes (1968 film)): chimpanzees: animal psychologist Zira (Kim Hunter) and surgeon Galen (Wright King). While unable to speak as his throat wound is healing, called "Bright Eyes" by Zira and placed with one of the captive primitive humans he later names "Nova", Taylor observes the enhanced society of talking apes and in a strict caste system: the gorillas being the military police, hunters and workers; the orangutans overseeing the affairs of government, science, and religion; and intellectual chimpanzees being mostly scientists. While their society is a theocracy similar to the beginnings of the human Industrial Era, the apes consider the primitive humans as\\
Document [3](Title: Planet of the Apes (1968 film)): Planet of the Apes (1968 film) Planet of the Apes is a 1968 American science fiction film directed by Franklin J. Schaffner. It stars Charlton Heston, Roddy McDowall, Kim Hunter, Maurice Evans, James Whitmore, James Daly and Linda Harrison. The screenplay by Michael Wilson and Rod Serling was loosely based on the 1963 French novel "La Planète des Singes" by Pierre Boulle. Jerry Goldsmith composed the groundbreaking avant-garde score. It was the first in a series of five films made between 1968 and 1973, all produced by Arthur P. Jacobs and released by 20th Century Fox. The film tells the\\
\\
Answer: In the 1968 film Planet of the Apes, Galen was played by Wright King. And in the tv series Planet of the Apes, Galen was played by Roddy McDowall.\\
\\
\\
Instruction: Write an accurate, engaging, and concise answer for the given question, possibly using the provided search results (some of which might be irrelevant).\\
\\
Question: Who has the highest goals in world football?\\
\\
Document [1](Title: Argentina–Brazil football rivalry): "Football Player of the Century", by IFFHS International Federation of Football History and Statistics, 1999, "South America Football Player of the Century", by IFFHS International Federation of Football History and Statistics. Pelé's 1281 goals are recognized by FIFA as the highest total achieved by a professional footballer, although the Soccer Statistic Foundation (rssf) recognizes only 767 goals in official mode, occupying the third place after Josef Bican (805) and Romario (772). For his part, Maradona has been named the best soccer player in World Cup history both by The Times and FourFourTwo, publication that also rewarded him as the "Best\\
Document [2](Title: Godfrey Chitalu): have beaten Gerd Müller's record of 85 goals in a year, the Football Association of Zambia claimed that the world record actually pertained to Godfrey Chitalu who had scored 116 goals (possibly 117) during the 1972 calendar year and 107 during the 1972 season. The difference of goals is due to first 9 goals being scored before the season officially started. The Football Association of Zambia presented the evidence to FIFA but a spokesperson responded that they would ratify neither Lionel Messi's nor Chitalu's records as they do not keep statistical track of domestic competitions. Nonetheless, it could constitute the\\
Document [3](Title: Godfrey Chitalu): highest official tally claimed by a national football association. Chitalu made his international debut on 29 June 1968 in a friendly match against Uganda in Lusaka which Zambia won 2–1. He scored his first goal in a 2–2 draw against the same team five days later. Chitalu played a prominent role during the World Cup qualification matches against Sudan with Zambia being eliminated on a strange rule which was peculiar to Africa and favoured the team that won the second leg. Despite the aggregate score being tied at 6–6 after Zambia won the first leg 4–2 and lost the return\\
\\
Answer:\\

\bottomrule[0.1em]
    \end{tabularx}
    }
    \caption{\textbf{Example of prompt} with one exemplar of question-documents-answer triple, used for ASQA.}
    \label{table:prompts}
\end{table*}

\end{document}

%% file: background.tex
Before describing our experimental setup, we more precisely define hallucination.
Following previous work, we define a \emph{hallucination} as text that is not grounded in the data provided to the model at either training or inference time~\cite{ji2023survey, agrawal2023language}.
Such hallucinations are sometimes characterized as \emph{open-domain} hallucinations in order to distinguish them from semantic deviations of the generated output in, e.g.,  machine translation---referred to as \emph{closed-domain} hallucination~\cite{ji2023survey}. 
It is important to distinguish \emph{factuality} from hallucination.
Specifically, factuality, or the factual correctness of (generated) text, refers to the quality of being based on a fact, i.e., world knowledge \cite{maynez-etal-2020-faithfulness}. 
Note that a model might output text that is grounded in its pre-training or inference-time data, yet is factually incorrect. 
While the number of grounded, factual errors may be reduced by improving the factuality of the data, preventing a model from generating text that is neither grounded nor factually accurate is a challenging problem with no known solution.